\documentclass[letterpaper,10pt,conference]{ieeeconf}

\usepackage{amsmath,amsfonts}
\usepackage{algorithm}
\usepackage{array}
\usepackage{subfig}
\usepackage{textcomp}
\usepackage{stfloats}
\usepackage{url}
\usepackage{verbatim}
\usepackage{graphicx}
\usepackage{cite}
\usepackage{amssymb}
\usepackage{hyperref}
\usepackage{multirow}
\usepackage{xcolor}
\usepackage{subcaption}
\usepackage{rotating}
\usepackage{pdflscape}
\usepackage{lettrine}
\usepackage{algpseudocode}
\usepackage{capt-of}
\usepackage{booktabs}
\usepackage{censor}
\usepackage{comment}

\begin{document}

\title{EndoNav: Semantic-to-Geometric Grounding for Language-Guided Robotic Endoscopic Examination}

\author{
Jecia Z. Y. Mao, Hisashi Ishida, Kathryn Jung, Masaru Ishii, Russell H. Taylor, Manish Sahu
}

\maketitle

\markboth{IEEE Transactions}
{Shell \MakeLowercase{\textit{et al.}}: A Sample Article Using IEEEtran.cls for IEEE Journals}

\begin{abstract}
Minimally invasive procedures performed within confined anatomical spaces depend on continuous endoscopic visualization. 
Current robotic endoscope systems can stabilize or reposition an endoscope, but they do not possess relevant context to provide effective visualization assistance.
We present EndoNav, an anatomy-grounded natural-language framework that translates high-level surgeon commands into autonomous endoscopic visualization behaviors within patient-specific sinonasal anatomy. Spoken surgeon commands are transcribed and interpreted by an endoscopic viewpoint agent conditioned on a patient-specific anatomical scene representation. Rather than generating robot motion directly, the viewpoint agent generates structured visualization objectives that are converted into target viewpoints and inspection trajectories, which are then executed through geometry-constrained endoscope motion planning and joint-space control. We evaluate EndoNav using a structured three-pass sinus examination across three CT-derived anatomical models. For one cadaveric specimen, autonomous visualization is compared with sinus examinations performed by two resident surgeons. EndoNav achieved mean visualization IoUs of $87.04\%$ and $84.37\%$ relative to the two surgeon examinations, compared with an inter-surgeon IoU of $87.44\%$, while recovering $92.91\%$ and $93.20\%$ of surgeon-observed anatomical surfaces, respectively. These results demonstrate the feasibility of grounding high-level anatomical commands into patient-specific geometric objectives and translating them into anatomically constrained robotic visualization behaviors.
\end{abstract}

\section{Introduction}

\lettrine[lines=2]{E}{endoscopic} visualization is fundamental to minimally invasive procedures performed through confined anatomical corridors. 
During an examination or procedure, the endoscope must be repeatedly repositioned to inspect anatomical structures, maintain spatial awareness, and provide clinically meaningful views of the operative field. Effective endoscope manipulation requires continuous coordination between patient-specific anatomical knowledge, the visualization objective, and endoscope control. Consequently, endoscope control imposes a substantial burden on the clinician holding the scope.

Robotic endoscope systems offer a means to reduce this burden through camera positioning and autonomous visualization assistance~\cite{taylor2016medical,yang2017medical,attanasio2021autonomy}. 
Existing systems have explored autonomous camera adjustment, image-guided navigation, and autonomous endoscope motion~\cite{pandya2014review,wang2023robotic,salcudean2022robot}. 
In endonasal procedures, where navigation occurs through narrow patient-specific corridors, anatomical models derived from preoperative imaging can additionally constrain motion planning~\cite{he2020endoscopic}. These approaches demonstrate that robotic systems can autonomously generate motion once a geometric navigation objective is specified. However, the surgeon generally remains responsible for translating the desired anatomical view into a target, trajectory, or low-level robot command.


\begin{figure}[H]
    \centering
    \includegraphics[width=1\linewidth]{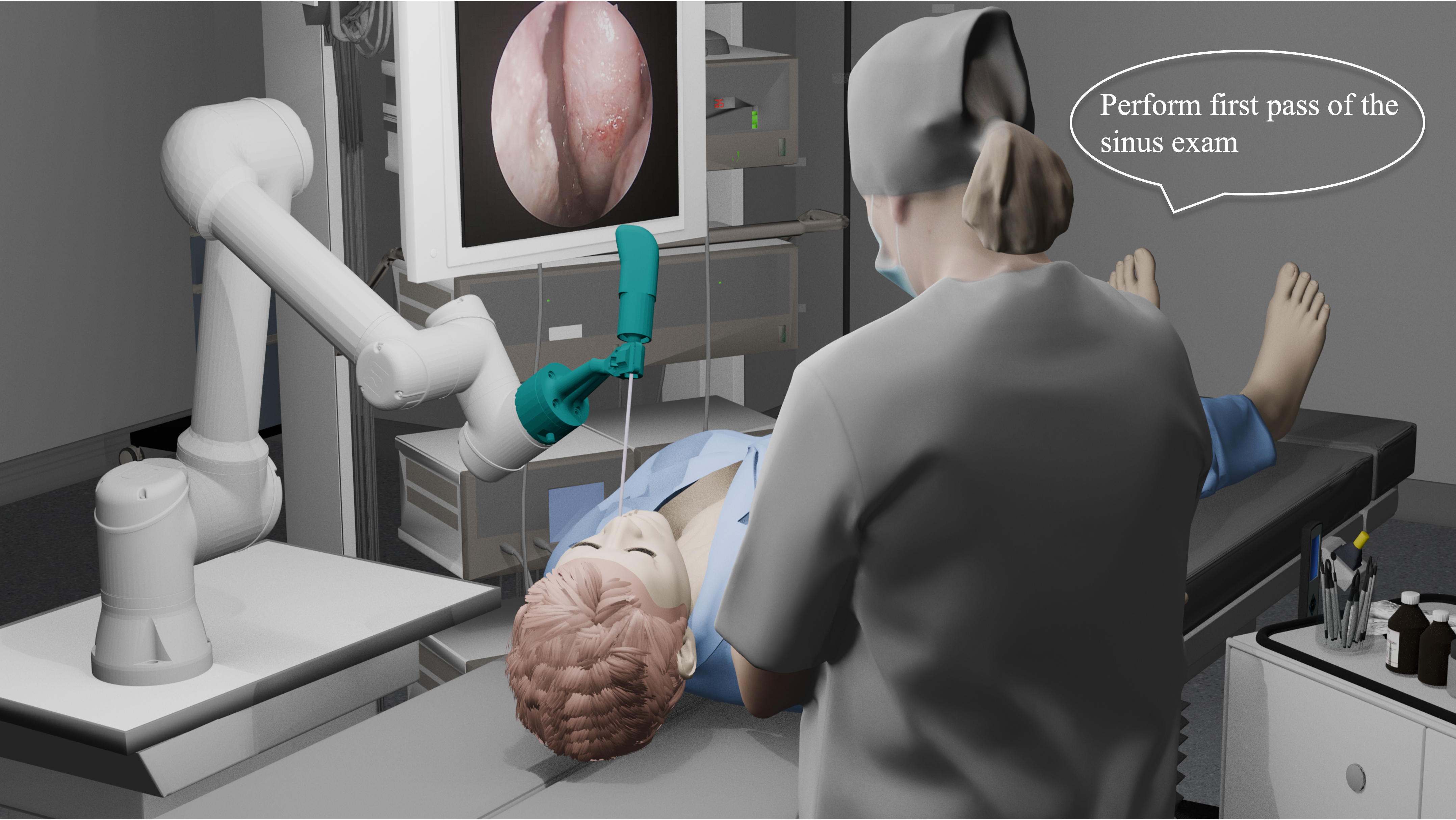}
    \caption{Conceptual illustration of the proposed language-guided autonomous sinus examination framework. A surgeon provides high-level verbal examination commands while a robot-held endoscope autonomously navigates within the patient-specific sinonasal anatomy to obtain the requested views.}
    \label{fig:placeholder}
\end{figure}

Natural-language interaction offers an intuitive way for communicating such objectives. Recent advances in large language models (LLMs) have enabled robots to interpret high-level instructions for task selection, policy composition, embodied reasoning, and navigation~\cite{brohan2023can,liang2023code,driess2023palme,shah2023lmnav}. This paradigm is increasingly being investigated in medical robotics, where LLMs can provide a communication and reasoning layer between clinicians and robotic systems~\cite{wang2026large,chen2026ai}. SuFIA~\cite{moghani2024sufia} demonstrated language-guided task orchestration capabilities for enabling surgical assistance, while \cite{killeen2024take} showed that conversational commands can specify clinically meaningful anatomical views for an intelligent robotic X-ray imaging system. Collectively, these studies establish natural language as a promising interface for specifying robotic and imaging objectives in clinical environments.

However, specifying an objective in clinical language and physically realizing it are distinct problems. An instruction such as ``inspect the Eustachian tube'' does not specify an endoscope position, orientation, or trajectory. Its geometric interpretation depends on the location and orientation of the requested structure within the patient's anatomy, the accessible free space, the current endoscope state, and the viewpoints from which the structure can be visualized. Existing language-guided medical robotic systems primarily map user intent to predefined actions, task-level plans, or imaging commands. They do not address this intermediate spatial reasoning problem: determining where a robot-held endoscope should be positioned and oriented within patient-specific anatomy to realize a semantically specified anatomical view. Recent reviews of LLM-enabled medical robotics similarly identify reconciliation of high-level semantic directives with precise, verifiable geometric actions as a key challenge toward embodied medical intelligence~\cite{wang2026large,chen2026ai}.


\begin{figure*}
    \centering
    \includegraphics[width=0.9\linewidth]{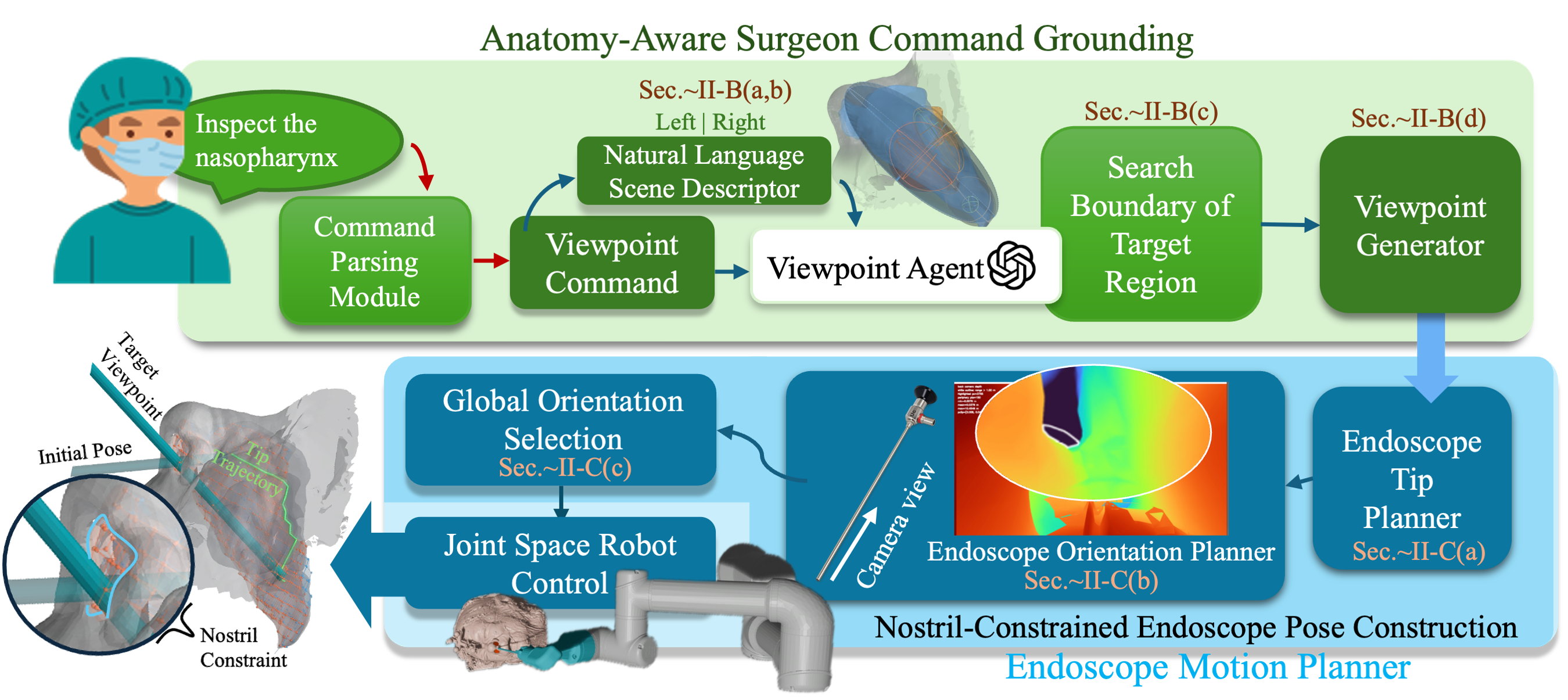}
    \caption{Overview of the autonomous sinus examination framework. Natural-language surgeon commands are grounded to patient-specific anatomical targets and converted into target viewpoints. A geometry-constrained motion planner generates anatomy-aware trajectories, constructs nostril-constrained endoscope poses, selects globally consistent orientations, and converts the resulting pose sequence into executable robot joint-space motion.}
    \label{fig:overview}
\end{figure*}

This challenge motivates a \emph{semantic-to-geometric grounding} architecture. Rather than allowing a foundation model to directly generate continuous robot motion, semantic reasoning and physical execution are explicitly separated. The language model produces a structured and interpretable anatomical visualization objective, while patient-specific geometry determines where the endoscope can be positioned and how the requested view can be physically achieved. This separation is particularly important in medical robotics, where stochastic model outputs, limited interpretability, and foundation-model inference latency complicate their direct use within low-level robot control loops~\cite{wang2026large}. Intermediate representations, including the anatomical target, spatial search region, target viewpoint, and robot trajectory, can instead be independently inspected and constrained before execution, while conventional geometric planning and joint-space control remain responsible for robot motion.


In this work, we present \textit{EndoNav}, a framework for \emph{patient-specific semantic-to-geometric grounding for surgeon-directed embodied visualization}. EndoNav operates within a simulation environment containing a CT-derived patient-specific anatomical model and a robot-held endoscope. The anatomical model provides a geometric representation of the nasal workspace and target structures. An endoscopic viewpoint agent interprets surgeon commands within this representation and produces structured spatial visualization objectives. The objective is subsequently grounded into a patient-specific endoscope viewpoint. The resulting viewpoint is realized through nostril-constrained endoscope motion planning, and joint-space robot control. 
EndoNav therefore bridges three representations that are typically treated separately: the anatomical language used by the surgeon, the patient-specific geometry required for spatial reasoning, and the robot motion required for embodied visualization.

We evaluate the proposed framework using a structured three-pass sinus examination as a representative visualization task. 
Autonomous examinations are performed across three CT-derived anatomical models. 
To assess the quality of the generated examinations, we conducted a cadaver study, where two resident surgeons performed a structured sinus examination on a specimen head. We compare the anatomical surfaces visualized during the EndoNav and surgeon examinations to evaluate whether distinct navigation behaviors accomplish comparable visualization objectives.

The main contributions of this work are:

\begin{itemize}
    \item an anatomy-grounded natural-language interface that translates clinically meaningful surgeon commands into structured visualization objectives that are expressed relative to patient-specific anatomy.
    \item a semantic-to-geometric grounding framework that transforms anatomy-level visualization requests into patient-specific endoscope viewpoints and robot-executable trajectories while separating probabilistic language reasoning from geometry-constrained motion generation; and
    \item an evaluation of autonomous sinus examination across three anatomical models, including quantitative comparison with examinations performed by two surgeons, to demonstrate EndoNav--surgeon visualization agreement of similar magnitude to the inter-surgeon agreement observed in the experimental specimen.
\end{itemize}

\section{Methodology}

\subsection{System Overview}

The proposed framework converts high-level surgeon commands in anatomical language into executable endoscope trajectories within patient-specific sinonasal anatomy. Given an instruction, the system identifies the requested anatomical target or navigation direction and generates a structured visualization objective grounded in the nasal-cavity representation. The objective is converted into a feasible target viewpoint, after which a geometry-constrained motion planner computes a collision-free tip trajectory with nostril-constrained shaft
orientations for robot execution. The framework is implemented in AMBF~\cite{munawar2019ambf} using CT-derived anatomical models and a robot-held rigid endoscope with virtual camera parameters and aperture matched to the physical endoscope used in the cadaver experiments.

As illustrated in Fig.~\ref{fig:overview}, the framework consists of two tightly coupled components. The \textbf{anatomy-grounded command module} uses GPT-4.1 as the viewpoint agent to map surgeon language to patient-specific visualization objectives and target viewpoints. Following this module, the \textbf{robot motion planner} generates a collision-free tip trajectory with nostril-constrained shaft orientations connecting the current pose to the selected viewpoint. This decomposition enables the LLM to reason about \textit{where the endoscope tip should be positioned to best visualize the target structure}, while geometric planning determines \textit{how the endoscope can safely reach that viewpoint}.

\begin{figure*}
    \centering
    \includegraphics[width=0.83\linewidth]{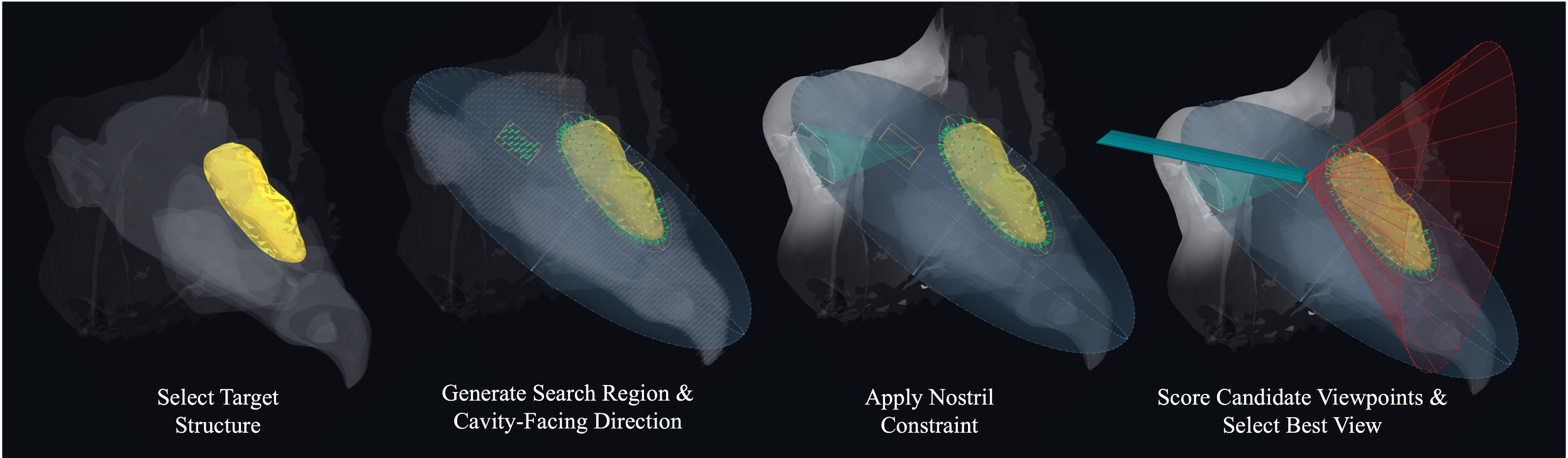}
    \caption{Anatomy-grounded target viewpoint generation. A target structure is localized within the patient-specific nasal frame,
and a bounded search region is generated with candidate viewpoints, each
consisting of a position and a nostril-constrained viewing orientation.
All candidates are then scored to select the final target viewpoint.}
    \label{fig:placeholder}
\end{figure*}

\subsection{Anatomy-Grounded Command Interpretation and Viewpoint Generation}

\paragraph{Patient-Specific Anatomical Representation}
Surgeon commands are naturally expressed relative to anatomy, whereas robot planning requires geometric targets. We therefore map segmented sinonasal anatomy into a compact patient-specific coordinate representation.

The left and right nasal cavities are treated as separate navigable
workspaces. For each cavity, we fit an ellipsoid to its segmented mesh
and represent it by its center $\boldsymbol{\mu}_{\mathrm{cavity}}$,
principal axes $\mathbf{A}_{\mathrm{cavity}}$, and semi-axis lengths
$\mathbf{r}_{\mathrm{cavity}}$:
\begin{equation}
\mathcal{F}_{\mathrm{cavity}}
=
\left(
\boldsymbol{\mu}_{\mathrm{cavity}},
\mathbf{A}_{\mathrm{cavity}},
\mathbf{r}_{\mathrm{cavity}}
\right).
\end{equation}
The principal axes are assigned the clinically meaningful directions
$\mathcal{K}=\{\mathrm{EN},\mathrm{LS},\mathrm{RF}\}$,
corresponding to entry--nasopharynx, lateral--septal, and roof--floor, respectively.

Each segmented structure $\mathcal{S}_i$ is processed to obtain the same
geometric representation
$\mathcal{F}_i=(\boldsymbol{\mu}_i,\mathbf{A}_i,\mathbf{r}_i)$
and expressed relative to its corresponding nasal-cavity frame. Together,
these representations form a patient-specific semantic scene graph,
\begin{equation}
\mathcal{G}
=
\{\mathcal{F}_i\}_{i\in\mathcal{I}_L}
\cup
\{\mathcal{F}_i\}_{i\in\mathcal{I}_R}
\cup
\{\mathcal{F}_i\}_{i\in\mathcal{I}_{\mathrm{shared}}},
\end{equation}
where $\mathcal{I}_L$, $\mathcal{I}_R$, and $\mathcal{I}_{\mathrm{shared}}$
denote the left-sided, right-sided, and shared anatomical structures between the two nasal passages, respectively.

\begin{figure}
    \centering
    \includegraphics[width=1\linewidth]{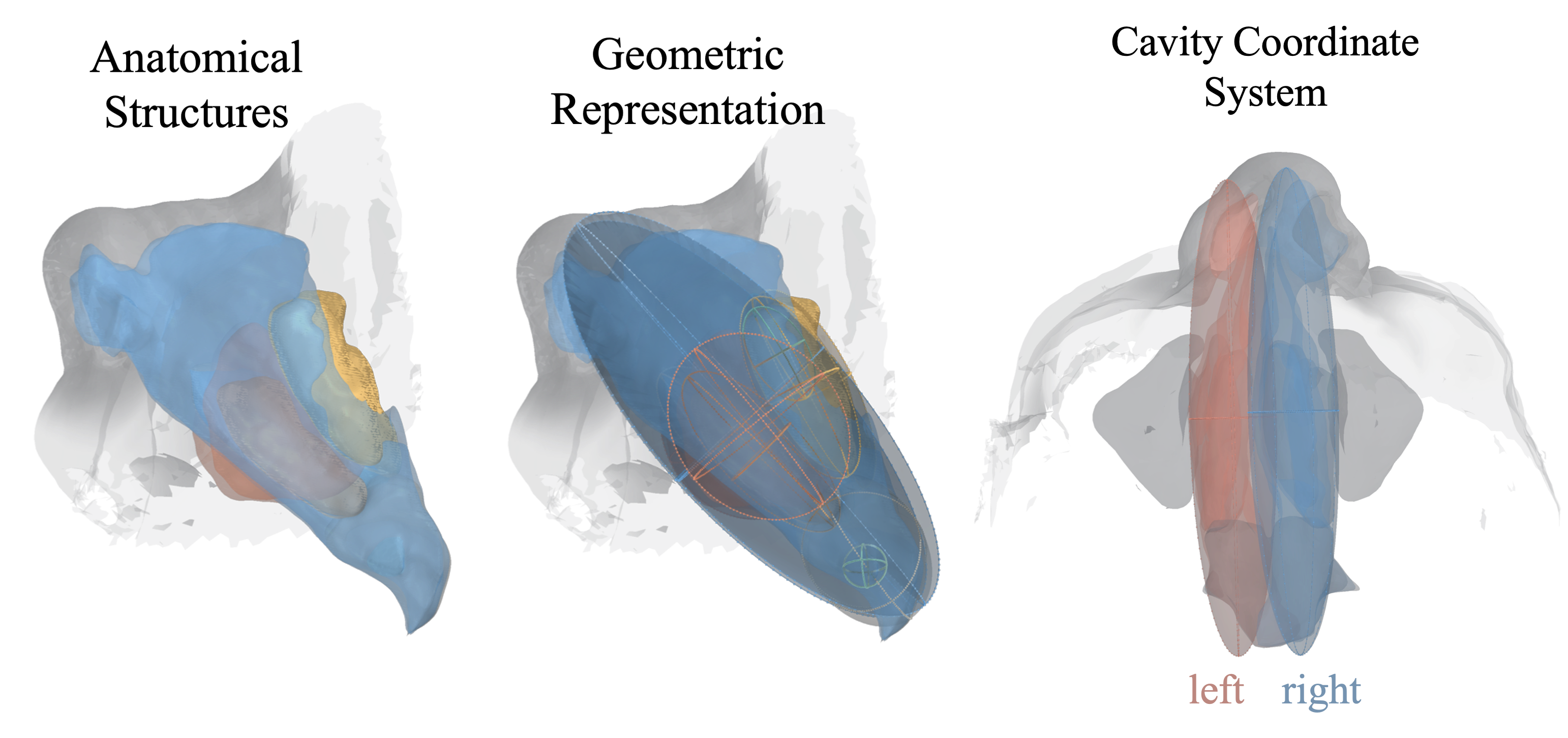}
    \caption{Construction of the anatomy-grounded 3-D representation.
Patient-specific anatomical structures are converted into compact geometric
representations and expressed relative to their corresponding nasal-cavity
coordinate frames to support scene description, command grounding, and
viewpoint planning.}
    \label{fig:placeholder}
\end{figure}

\paragraph{Scene Graph Descriptor}

The semantic description of each structure consists of three components: its
location within the corresponding nasal cavity, its cavity-facing direction,
and its longitudinal span.

For each structure $\mathcal{S}_i$, the surface point closest to the
corresponding cavity centroid is used to describe its anatomy-relative
location,
\begin{equation}
\mathbf{q}_i =
\arg\min_{\mathbf{x}\in\mathcal{S}_i}
\left\|
\mathbf{x}-\boldsymbol{\mu}_{\mathrm{cavity}}
\right\|_2,
\qquad
\boldsymbol{\delta}_i =
\mathbf{q}_i-\boldsymbol{\mu}_{\mathrm{cavity}}.
\end{equation}
The displacement $\boldsymbol{\delta}_i$ is projected onto the labeled cavity
axes and normalized by the corresponding cavity extents to obtain a controlled
semantic description of where the structure lies.

The cavity-facing direction is estimated from surface normals exposed to the
corresponding nasal cavity. Let $\mathbf{x}_m$ and $\mathbf{n}_m$ denote the
$m$th sampled surface point and its unit normal, respectively, and let
$\phi_c(\cdot)$ denote the SDF of the corresponding cavity. The exposed surface
set is defined as
\begin{equation}
\mathcal{E}_i =
\left\{
m \;\middle|\;
\exists \lambda>0,\;
\begin{aligned}
&\phi_c(\mathbf{x}_m+\lambda\mathbf{n}_m)\leq0 \\
&\vee\;
\phi_c(\mathbf{x}_m-\lambda\mathbf{n}_m)\leq0
\end{aligned}
\right\}.
\end{equation}
The normals in $\mathcal{E}_i$ are oriented toward the cavity and averaged to
obtain the cavity-facing direction $\mathbf{g}_i$, which is mapped to the most
closely aligned labeled cavity axis.

Finally, the longitudinal span is defined by the endpoints of the longest
principal axis of $\mathcal{F}_i$,
\begin{equation}
\ell_{\mathrm{span}}
=
\arg\max_{\ell} r_{i,\ell},
\qquad
\mathbf{x}_i^{\pm}
=
\boldsymbol{\mu}_i
\pm
r_{i,\ell_{\mathrm{span}}}
\mathbf{a}_{i,\ell_{\mathrm{span}}}.
\end{equation}
The endpoints are projected onto the entry--nasopharynx axis to describe how
the structure extends along the navigable corridor.

Together, these quantities provide a compact anatomy-relative description of
each structure. For example:
\textit{``The left middle turbinate lies superior and slightly lateral relative
to the left nasal-cavity center, faces the nasal septum, and extends from
slightly toward the entry side to considerably toward the nasopharynx.''}

\paragraph{Language-Guided Examination Goal Generation}
Given the surgeon command and the patient-specific scene description, the agent
identifies the referenced anatomical target or requested motion direction and
predicts an anatomy-relative region in which the endoscope tip should be
positioned. The prediction is expressed along the clinically meaningful cavity
directions $\mathcal{K}$ defined above.

For each direction $k\in\mathcal{K}$, the qualitative description is
deterministically mapped to a normalized interval,
\begin{equation}
d_k \longmapsto [l_k,u_k],
\end{equation}
where $l_k$ and $u_k$ denote the lower and upper bounds of the corresponding
magnitude bucket. These intervals are scaled by the patient-specific cavity
geometry to define the bounded 3-D tip search region,
\begin{equation}
\begin{aligned}
\mathcal{B}
&=
\left\{
\boldsymbol{\mu}_{\mathrm{cavity}}
+
\boldsymbol{\delta}
\;\middle|\;
\boldsymbol{\delta}^{\top}
\mathbf{a}_{\mathrm{cavity},k}
\in
[l_k r_{\mathrm{cavity},k},\,u_k r_{\mathrm{cavity},k}]
\right\},
\\
&\hspace{4.5cm}
k\in\mathcal{K}.
\end{aligned}
\end{equation}

\paragraph{Target Viewpoint Optimization}
We sample feasible positions within $\mathcal{B}$ and generate
nostril-pivot-constrained orientation proposals at each position. Each candidate
viewpoint is represented as
\begin{equation}
T_c = (\mathbf{p}_c,\mathbf{R}_c),
\end{equation}
where $\mathbf{p}_c$ is the candidate endoscope-tip position and
$\mathbf{R}_c$ its orientation. The resulting pose candidates are scored using
viewing-direction alignment and target framing with equal coefficients,
\begin{equation}
S(T_c)
=
\frac{1}{2}S_{\mathrm{align}}
+
\frac{1}{2}S_{\mathrm{framing}},
\end{equation}
where
\begin{equation}
S_{\mathrm{align}}
=
\hat{\mathbf{f}}_c^{\top}(-\mathbf{g}_i),
\qquad
S_{\mathrm{framing}}
=
\frac{1}{1+\Delta_x+\Delta_y}.
\end{equation}
Here, $\hat{\mathbf{f}}_c$ denotes the optical-forward direction induced by
$\mathbf{R}_c$, and $\mathbf{g}_i$ is the cavity-facing direction of target
structure $\mathcal{S}_i$ defined above. $\Delta_x$ and $\Delta_y$ denote the
normalized projected spans of the target along the camera image axes. The
alignment term ranges from $-1$ to $1$, while the framing term lies in
$(0,1]$ and favors a smaller target footprint, providing a wider contextual
view of the surrounding anatomy.

The final viewpoint is selected as
\begin{equation}
T_c^{\star}
=
\arg\max_{T_c\in\mathcal{T}_{\mathrm{feasible}}}
S(T_c).
\end{equation}

For directional commands without an explicit target structure, the user-specified
anatomical viewing direction is used directly to select among the feasible
orientation proposals and generate the target pose.

\subsection{Geometry-Constrained Endoscope Motion Planning and Execution}

\begin{figure}
    \centering
    \includegraphics[width=1\linewidth]{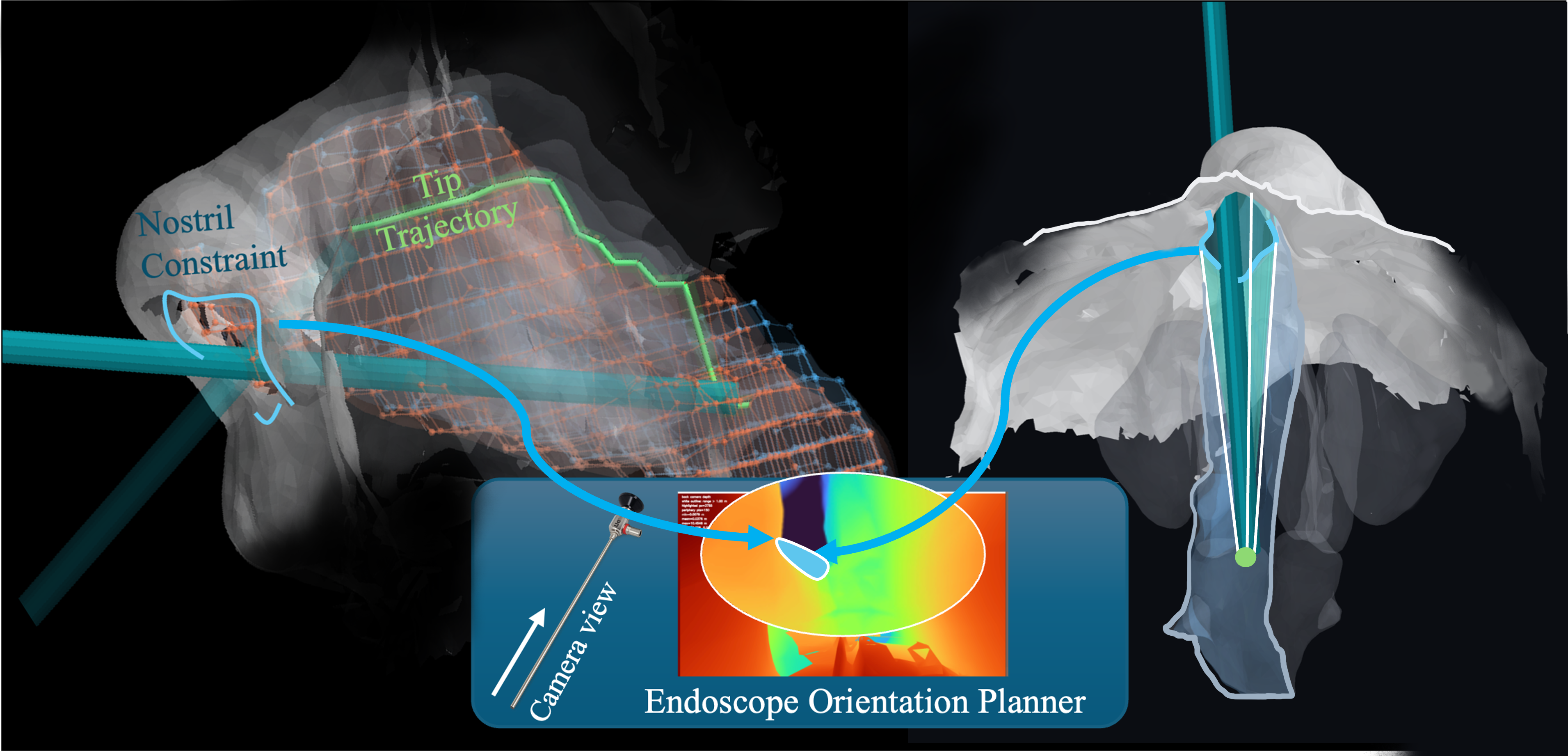}
\caption{Anatomy-constrained endoscope trajectory and nostril-constrained orientation generation. The planned tip trajectory remains within the patient-specific nasal free space. Because the endoscope is rigid and straight, its orientation at each tip waypoint is constrained by the nostril through which the shaft must pass. A virtual backward-facing depth camera observes only the surrounding anatomy and back-projects the visible nostril boundary to generate feasible shaft orientations. These candidates are subsequently used for global orientation selection along the trajectory.}
    \label{fig:placeholder}
\end{figure}

\paragraph{Collision-Aware Tip Trajectory Planning}
We construct a signed distance field (SDF) from the segmented anatomy and retain free-space voxels satisfying
\begin{equation}
\phi(\mathbf{x}) \ge d_{\min},
\end{equation}
where $d_{\min}$ is the required anatomical clearance. The retained voxels form a connected navigation graph. The current tip and target viewpoint are projected onto the graph, and A* computes a collision-free path following the graph-based formulation of~\cite{he2020endoscopic}. The path is shortcut-smoothed under the same clearance constraint and resampled into tip waypoints
\begin{equation}
\mathcal{W}=\{\mathbf{p}_0,\mathbf{p}_1,\ldots,\mathbf{p}_N\}.
\end{equation}

\paragraph{Nostril-Constrained Orientation Generation}
Because the endoscope is rigid and straight, its orientation at a fixed tip
position is constrained by the nostril through which the shaft passes. We
therefore treat the nostril as a geometric pivot constraint: for each waypoint
$\mathbf{p}_i$, feasible orientations must keep the endoscope shaft within the
nostril opening.

A backward-facing virtual depth camera at $\mathbf{p}_i$ observes the local
nasal boundary surrounding the shaft. The observed geometry is back-projected
to generate a set of feasible shaft orientations,
\begin{equation}
\mathcal{R}_i=\{R_{i,1},\ldots,R_{i,K_i}\}.
\end{equation}
Each candidate preserves the tip position while pivoting the shaft through the
available nostril geometry. Rotational continuity resolves the remaining roll
degree of freedom. If the nostril boundary is not visible, predefined
periphery-plane geometry is used instead.

\paragraph{Global Pose Optimization and Robot Execution}

Independent selection of locally feasible orientations may produce abrupt camera motion or orientations inconsistent with the desired terminal visualization. We therefore select the orientation sequence globally.

Each orientation candidate defines a camera viewing direction
$\mathbf{c}_{i,k}$. A baseline viewing-direction trajectory
$\{\bar{\mathbf{c}}_i\}_{i=0}^{N}$ is constructed between the initial camera
viewing direction and the target viewing direction generated by the viewpoint
optimizer. Dynamic programming selects the candidate index $k_i$ at each
waypoint by minimizing
\begin{equation}
C =
\sum_{i=1}^{N}
w_{\mathrm{loc}}
\theta(\mathbf{c}_{i-1,k_{i-1}},\mathbf{c}_{i,k_i})^2
+
\sum_{i=0}^{N}
w_{\mathrm{goal}}
\theta(\mathbf{c}_{i,k_i},\bar{\mathbf{c}}_{i})^2 ,
\end{equation}
where $\theta(\cdot,\cdot)$ denotes angular distance. The first term promotes
smooth endoscope rotation between consecutive waypoints, while the second
maintains consistency with the baseline viewing-direction trajectory.

The selected orientations and planned tip positions define the final endoscope pose trajectory,

\begin{equation}
\mathcal{T}^{\star}
=
\{T_0^{\star},\ldots,T_N^{\star}\},
\qquad
T_N^{\star}=T_c^{\star}.
\end{equation}

Inverse kinematics is solved sequentially along this trajectory using the previous joint configuration as initialization to promote joint-space continuity. The resulting robot trajectory is executed in simulation.


\section{Experiments and Results}

\begin{figure*}
\centering
\includegraphics[width=0.8\linewidth]{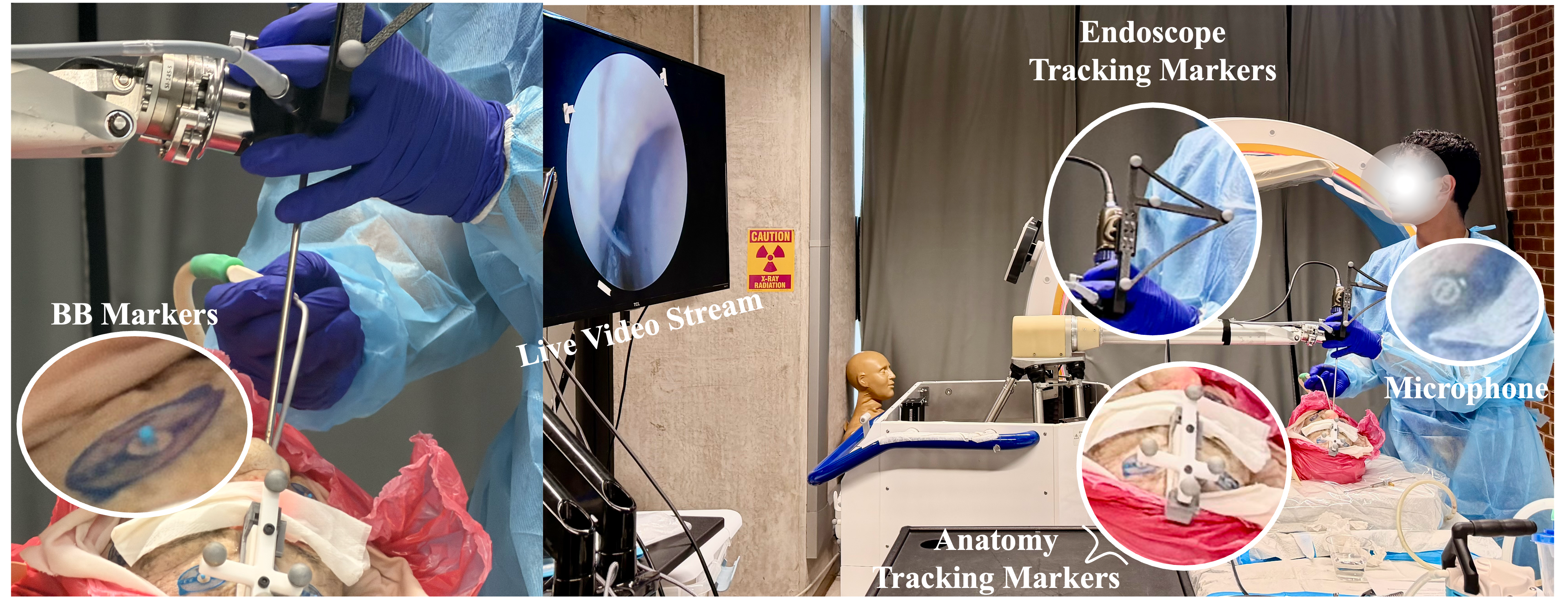}
\caption{Ex vivo validation setup. A surgeon manually operates the rigid endoscope while viewing the live endoscopic stream on an external display. Tracking markers are attached to the endoscope and cadaveric head reference, and BB fiducials are used to register the tracking frame to the anatomical frame. Surgeon narration is recorded synchronously to identify target arrival, inspection completion, and transitions between anatomical structures.}
\label{fig:experiment_setup}
\end{figure*}

\subsection{Experimental Setup}

The experimental setup evaluates whether EndoNav can autonomously execute a structured nasal endoscopy examination and reproduce the anatomical coverage achieved by resident surgeons. Reference examinations were acquired on an ex vivo cadaveric head using a rigid
$0^\circ$ endoscope mounted on the robot and manually manipulated by each resident
surgeon. An Atracsys optical tracking system tracked rigid marker frames attached
to the endoscope and cadaveric head, while BB fiducials provided registration
between the tracker and CT-derived anatomical frames. Synchronized endoscopic
video and surgeon narration were recorded to identify examination progression
and the anatomical structures being inspected. A volumetric CT scan of the
specimen was used to construct the anatomical models for subsequent evaluation.

To reconstruct each surgeon examination in the same anatomical frame used by EndoNav, the physical endoscope, optical tracker, and CT-derived anatomy were registered into a common coordinate system. Camera calibration provided the endoscope intrinsics and effective circular aperture, hand--eye calibration estimated ${}^{\mathcal{M}}\mathbf{T}_{\mathcal{C}}$, and BB fiducials were used to compute the tracker-to-anatomy transformation ${}^{\mathcal{A}}\mathbf{T}_{\mathcal{T}}$. The camera pose was therefore recovered as
\begin{equation}
{}^{\mathcal{A}}\mathbf{T}_{\mathcal{C}}(t)
=
{}^{\mathcal{A}}\mathbf{T}_{\mathcal{T}}
{}^{\mathcal{T}}\mathbf{T}_{\mathcal{M}}(t)
{}^{\mathcal{M}}\mathbf{T}_{\mathcal{C}} .
\label{eq:camera_pose_anatomy}
\end{equation}
These poses and calibrated camera parameters were used to instantiate the virtual endoscope in AMBF ~\cite{munawar2019ambf}, aligning the simulated view with the recorded physical examination.

\subsection{Validation Procedure}

We evaluate whether EndoNav can autonomously execute a structured nasal endoscopy examination and reproduce the anatomical visual coverage achieved during manual examination. For each nostril, the examination was organized into three passes: an overall nasal-cavity inspection including the floor, septum, and roof; inspection of the middle turbinate and middle-meatus region; and inspection of the posterior nasal cavity including the nasopharyngeal and Eustachian-tube regions. Surgeons manually performed this protocol while synchronized endoscopic video, endoscope tracking, anatomy-reference tracking, and narration were recorded. The narration established the ordered anatomical objectives for each pass. The same examination sequence was converted into natural-language instructions and provided to EndoNav, which independently grounded each objective in the patient-specific anatomy, generated the corresponding target viewpoint, and planned the complete entry--examination--exit trajectory for each pass.

To compare visualized anatomy, point clouds generated by the simulator from the endoscopic observations were first registered to the CT-derived anatomy using fiducial correspondences and then refined with iterative closest point (ICP) alignment. The accumulated registered point clouds were associated with the corresponding segmented anatomical surfaces to produce structure-specific visualization maps. This enabled surgeon and EndoNav coverage to be compared directly on the same anatomical model.

For each target structure $j$, let $S^j$ denote its segmented surface, with surgeon-observed and robot-observed subsets $S_{\mathrm{surg}}^j$ and $S_{\mathrm{rob}}^j$. Agreement was quantified using
\begin{equation}
\mathrm{IoU}^{j} =
\frac{|S_{\mathrm{surg}}^{j}\cap S_{\mathrm{rob}}^{j}|}
{|S_{\mathrm{surg}}^{j}\cup S_{\mathrm{rob}}^{j}|},
\qquad
\mathrm{Recall}_{\mathrm{surg}}^{j} =
\frac{|S_{\mathrm{surg}}^{j}\cap S_{\mathrm{rob}}^{j}|}
{|S_{\mathrm{surg}}^{j}|}.
\end{equation}
Metrics were computed independently for each structure and nasal side, and inter-surgeon agreement was computed using the same formulation. Soft visual coverage for structure $k$ was computed as
\begin{equation}
C_k =
\frac{100}{|\mathcal{X}_k|}
\sum_{\mathbf{x}\in\mathcal{X}_k}
\exp\left[
-\frac{1}{2}
\left(
\frac{
\min_{\mathbf{p}\in\mathcal{P}_{\mathrm{ICP}}}
\|\mathbf{x}-\mathbf{p}\|_2
}{\sigma}
\right)^2
\right],
\end{equation}
where $\mathcal{X}_k$ denotes sampled surface points of structure $k$, $\mathcal{P}_{\mathrm{ICP}}$ is the registered point cloud, and $\sigma$ controls the spatial falloff.

\subsection{Results}

\vspace{3mm}

\begin{figure}[t]
    \centering
    \includegraphics[width=0.95\linewidth]{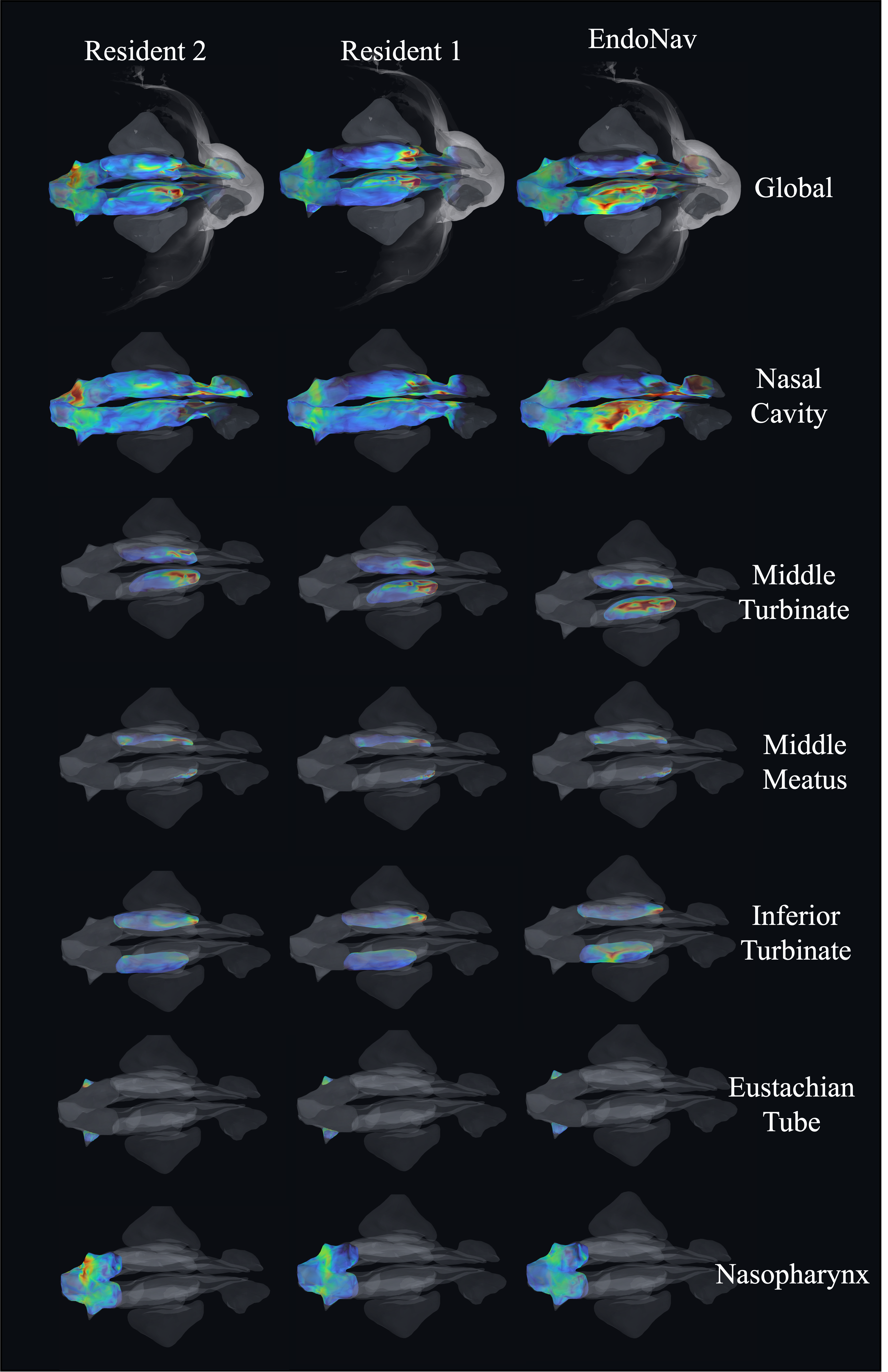}
\caption{Comparison of anatomical visualization achieved by EndoNav and two
resident examinations. Surface heatmaps show accumulated visualization
globally and across individual anatomical structures. Image-plane centerness
assigns higher weight to observations near the image center, while
depth preference favors nearer observations. The accumulated weights are
mapped onto the registered anatomical surface, such that intensity reflects
both observation preference and correspondence to the surface. Red indicates
higher intensity and blue lower intensity.}
    \label{fig:validation}
\end{figure}

\begin{table}[t]
\centering
\caption{Structure-level comparison between EndoNav and examinations performed by two
surgeons in the experimental specimen. Coverage (Cov.),
intersection-over-union (IoU), and recall (Rec.) are reported as percentages
for the left (L) and right (R) nasal passages. EndoNav--R1 and EndoNav--R2 denote
agreement between the autonomous examination and each resident examination,
while R1--R2 denotes inter-resident agreement. Summary IoU and recall values
are reported as macro mean $\pm$ standard deviation across all structures and
both nasal passages.}
\label{tab:coverage_comparison_bilateral}

\footnotesize
\setlength{\tabcolsep}{2.3pt}
\renewcommand{\arraystretch}{1.08}

\resizebox{\columnwidth}{!}{%
\begin{tabular}{@{}ll cc ccc cc@{}}
\toprule

& &
\multicolumn{2}{c}{\textbf{Coverage}}
& \multicolumn{3}{c}{\textbf{IoU}}
& \multicolumn{2}{c}{\textbf{Recall}} \\

\cmidrule(lr){3-4}
\cmidrule(lr){5-7}
\cmidrule(lr){8-9}

\textbf{Structure}
& \textbf{Side}
& \textbf{R1}
& \textbf{R2}
& \multicolumn{2}{c}{\textbf{EndoNav}}
& \textbf{R1--R2}
& \multicolumn{2}{c}{\textbf{EndoNav}} \\

\cmidrule(lr){5-6}
\cmidrule(lr){8-9}

& &
& &
\textbf{R1}
& \textbf{R2}
&
& \textbf{R1}
& \textbf{R2} \\

\midrule

\multirow{2}{*}{Nasal cavity}
& L & 63.02 & 73.37 & 80.35 & 79.42 & 78.75 & 92.43 & 85.06 \\
& R & 74.16 & 70.02 & 86.68 & 85.97 & 90.31 & 92.30 & 94.23 \\

\addlinespace[1pt]

\multirow{2}{*}{Middle turbinate}
& L & 46.61 & 35.24 & 74.17 & 71.01 & 74.84 & 82.29 & 92.67 \\
& R & 65.27 & 58.42 & 92.51 & 86.69 & 87.35 & 96.24 & 98.87 \\

\addlinespace[1pt]

\multirow{2}{*}{Middle meatus}
& L & 62.58 & 53.91 & 88.99 & 79.38 & 84.99 & 97.59 & 99.71 \\
& R & 51.05 & 43.82 & 86.32 & 82.97 & 82.46 & 91.68 & 97.13 \\

\addlinespace[1pt]

\multirow{2}{*}{Inferior turbinate}
& L & 59.46 & 66.84 & 87.31 & 77.78 & 82.86 & 90.32 & 80.05 \\
& R & 62.37 & 58.91 & 83.77 & 85.68 & 86.83 & 92.76 & 96.41 \\

\addlinespace[1pt]

\multirow{2}{*}{Eustachian tube}
& L & 95.35 & 95.15 & 91.16 & 91.19 & 99.33 & 94.10 & 94.26 \\
& R & 73.55 & 75.58 & 94.01 & 95.91 & 97.40 & 99.12 & 98.90 \\

\addlinespace[1pt]

\multirow{2}{*}{Nasopharynx}
& L & 60.52 & 63.80 & 86.57 & 86.15 & 90.78 & 90.48 & 87.85 \\
& R & 66.71 & 68.23 & 92.66 & 90.32 & 93.40 & 95.63 & 93.22 \\

\midrule

\multicolumn{2}{l}{\textbf{Macro mean}}
& \multicolumn{2}{c}{--}
& \textbf{87.04}
& \textbf{84.37}
& \textbf{87.44}
& \textbf{92.91}
& \textbf{93.20} \\

\multicolumn{2}{l}{\textbf{SD}}
& \multicolumn{2}{c}{--}
& 5.66
& 6.75
& 7.27
& 4.35
& 6.05 \\

\bottomrule
\end{tabular}%
}

\end{table}

\textbf{1) Agreement with surgeon visual coverage:}
Figure~\ref{fig:validation} qualitatively compares the anatomical regions visualized by EndoNav with those observed during the two surgeon examinations. The accumulated surface heatmaps show similar visualization patterns across the nasal cavity and the individual target structures, indicating that the autonomous examination recovered substantial portions of the anatomy observed during manual examination.

Table~\ref{tab:coverage_comparison_bilateral} quantifies this agreement. Across all evaluated structures and both nasal passages, EndoNav achieved a mean IoU of $(87.04 \pm 5.66)\%$ relative to R1 and $(84.37 \pm 6.75)\%$ relative to R2. The corresponding inter-surgeon IoU was $(87.44 \pm 7.27)\%$. Thus, the overall EndoNav--surgeon visualization agreement was comparable to the variability observed between the two manual examinations.

EndoNav also recovered a large proportion of the anatomical surfaces observed by each surgeon, with macro mean recall values of $(92.91 \pm 4.35)\%$ and $(93.20 \pm 6.05)\%$ relative to R1 and R2, respectively. Structure-level agreement remained high for most targets. For example, EndoNav achieved IoUs of $91.16$--$95.91\%$ for the Eustachian tube and $86.15$--$92.66\%$ for the nasopharynx across the two surgeons and nasal sides. The largest discrepancy occurred for the left middle turbinate relative to R2, where the IoU decreased to $71.01\%$, indicating that agreement remained dependent on the specific anatomical region and surgeon examination.
\begin{table}[t]
\centering
\caption{Structure-level autonomous examination coverage across three anatomical specimens. All values are reported as percentages. Mean coverage is reported as mean $\pm$ standard deviation across the left and right passages of all three specimens.}
\label{tab:cross_anatomy_coverage}

\scriptsize
\setlength{\tabcolsep}{2.7pt}
\renewcommand{\arraystretch}{1.08}

\begin{tabular}{@{}l cc cc cc c@{}}
\toprule
& \multicolumn{2}{c}{\textbf{Exp. Head}}
& \multicolumn{2}{c}{\textbf{Head 2}}
& \multicolumn{2}{c}{\textbf{Head 3}}
& \textbf{Mean} \\
\cmidrule(lr){2-3}
\cmidrule(lr){4-5}
\cmidrule(lr){6-7}

\textbf{Structure}
& \textbf{L} & \textbf{R}
& \textbf{L} & \textbf{R}
& \textbf{L} & \textbf{R}
& \textbf{EndoNav} \\
\midrule

Nasal cavity
& 67.67 & 72.99
& 71.83 & 73.66
& 68.33 & 76.19
& \textbf{71.78 $\pm$ 3.26} \\

Middle turbinate
& 43.43 & 65.70
& 64.06 & 79.43
& 63.80 & 76.26
& \textbf{65.45 $\pm$ 12.66} \\

Middle meatus
& 67.35 & 50.01
& 49.16 & 62.32
& 67.20 & 76.26
& \textbf{62.05 $\pm$ 10.66} \\

Inferior turbinate
& 55.61 & 64.37
& 71.09 & 83.80
& 54.84 & 60.27
& \textbf{65.00 $\pm$ 11.00} \\

Eustachian tube
& 92.84 & 77.01
& 73.17 & 67.77
& 48.01 & 44.00
& \textbf{67.13 $\pm$ 18.42} \\

Nasopharynx
& 57.40 & 65.86
& 67.58 & 63.44
& 52.89 & 58.55
& \textbf{60.95 $\pm$ 5.61} \\

\bottomrule
\end{tabular}

\end{table}

\textbf{2) Autonomous examination across patient-specific anatomies:}
Table~\ref{tab:cross_anatomy_coverage} summarizes autonomous visualization coverage across three CT-derived anatomical specimens. EndoNav successfully generated examination trajectories for both nasal passages of all three specimens, while the amount of observed target surface varied across anatomical structures. Mean coverage was highest for the nasal cavity $(71.78 \pm 3.26)\%$, followed by the Eustachian tube $(67.13 \pm 18.42)\%$, middle turbinate $(65.45 \pm 12.66)\%$, inferior turbinate $(65.00 \pm 11.00)\%$, and middle meatus $(62.05 \pm 10.66)\%$. Nasopharyngeal coverage was $(60.95 \pm 5.61)\%$. These results demonstrate that the same anatomy-grounded command and planning framework can generate visualization behaviors across different patient-specific geometries, while also revealing anatomy- and structure-dependent variation in achievable coverage.


\begin{table}[t]
\centering
\caption{Examination execution and global motion planning times across three
anatomical specimens. EndoNav values are reported as execution/planning time.
All values are in seconds.}
\label{tab:robot_timing_all_cases}

\scriptsize
\setlength{\tabcolsep}{2.6pt}
\renewcommand{\arraystretch}{1.10}

\resizebox{\columnwidth}{!}{%
\begin{tabular}{ll ccc cc cc}
\toprule

\textbf{Pass}
& \textbf{Side}
& \multicolumn{3}{c}{\textbf{Experiment Head}}
& \textbf{Head 2}
& \textbf{Head 3}
& \multicolumn{2}{c}{\textbf{Mean}} \\

\cmidrule(lr){3-5}
\cmidrule(lr){8-9}

&
& \textbf{R1}
& \textbf{R2}
& \textbf{EndoNav}
& \textbf{EndoNav}
& \textbf{EndoNav}
& \textbf{EndoNav}
& \textbf{Resident} \\

&
& \textbf{Exec.}
& \textbf{Exec.}
& \textbf{Exec./Plan.}
& \textbf{Exec./Plan.}
& \textbf{Exec./Plan.}
& \textbf{Exec./Plan.}
& \textbf{Exec.} \\

\midrule

\multirow{2}{*}{Pass 1}
& L & 65 & 54 & 105 / 22 & 39 / 23 & 38 / 14
& \multirow{2}{*}{70 / 17}
& \multirow{2}{*}{59} \\
& R & 77 & 39 & 89 / 19 & 73 / 11 & 74 / 16
& & \\

\addlinespace[1pt]

\multirow{2}{*}{Pass 2}
& L & 56 & 155 & 103 / 17 & 53 / 15 & 31 / 11
& \multirow{2}{*}{62 / 14}
& \multirow{2}{*}{86} \\
& R & 72 & 59 & 62 / 10 & 56 / 11 & 68 / 18
& & \\

\addlinespace[1pt]

\multirow{2}{*}{Pass 3}
& L & -- & 130 & 90 / 17 & 88 / 33 & 60 / 22
& \multirow{2}{*}{74 / 23}
& \multirow{2}{*}{104} \\
& R & 110 & 72 & 71 / 16 & 64 / 27 & 72 / 26
& & \\

\midrule

\multirow{2}{*}{\textbf{Total}}
& \textbf{L}
& \textbf{121}
& \textbf{339}
& \textbf{298 / 56}
& \textbf{180 / 70}
& \textbf{129 / 47}
& \multirow{2}{*}{\textbf{206 / 55}}
& \multirow{2}{*}{\textbf{256}} \\

& \textbf{R}
& \textbf{259}
& \textbf{170}
& \textbf{222 / 45}
& \textbf{193 / 49}
& \textbf{214 / 60}
& & \\

\bottomrule
\end{tabular}%
}

\vspace{1mm}

\begin{minipage}{\columnwidth}
\scriptsize
\textit{Note:}
R1 did not complete Pass 3 on the left side because the narrow nasal passage prevented posterior access; therefore, the reported R1 left-side total includes only Passes 1 and 2. Resident mean times were computed from completed measurements, with the incomplete R1 left-side examination excluded from the mean total.
\end{minipage}

\end{table}

\textbf{3) Examination and planning time:}
Execution times are reported in Table~\ref{tab:robot_timing_all_cases}. Across the three anatomical specimens, autonomous examination time varied by anatomical side and scoping pass, reflecting differences in planned trajectory length and accessibility of the requested structures. For the experimental specimen used for surgeon comparison, the autonomous examination required 298~s and 222~s for the left and right passages, respectively. These durations were of the same order as the corresponding manual examinations, although direct timing comparison is limited by differences in execution behavior and by the incomplete third left-side pass for R1.

The computational cost of global robot motion planning is summarized in Table~\ref{tab:robot_timing_all_cases}. Across all specimens and sides, mean planning times were 17~s, 14~s, and 23~s for the three examination passes, respectively, with a mean cumulative planning time of 55~s per nasal-side examination. These results indicate that high-level anatomical commands can be converted into complete endoscope trajectories on a time scale compatible with sequential, command-driven examination in the current simulation framework.

 \section{DISCUSSION}

This study investigated whether high-level surgeon intent expressed through anatomical language can be grounded to patient-specific anatomy and converted into autonomous endoscopic visualization behavior. In the experimental specimen, EndoNav achieved macro mean IoUs of $(87.04 \pm 5.66)\%$ and $(84.37 \pm 6.75)\%$ relative to the two surgeon examinations, compared with an inter-surgeon IoU of $(87.44 \pm 7.27)\%$. These results do not establish equivalence to surgeon performance, but show that autonomous--surgeon visualization agreement was comparable to the variability observed between the two manual examinations. Surgeon-surface recalls of $(92.91 \pm 4.35)\%$ and $(93.20 \pm 6.05)\%$ further indicate that EndoNav reproduced a substantial portion of the anatomy visualized by the surgeons.

These findings support evaluating autonomous endoscope behavior based on resulting anatomical visualization rather than direct trajectory similarity. The same target can be inspected from multiple valid poses and trajectories, and the two manual examinations themselves differed in coverage. Surface-based comparison therefore provides a more appropriate measure of whether different navigation behaviors achieve a similar visualization objective.

Evaluation across three CT-derived anatomical specimens showed that the same grounding and planning framework could generate bilateral examination trajectories without anatomy-specific modification, although coverage remained structure- and anatomy-dependent. The Eustachian tube showed the greatest inter-specimen variability, with a mean coverage of $(67.13 \pm 18.42)\%$. Its small size relative to the other evaluated structures makes the coverage metric more sensitive to minor segmentation discrepancies. More broadly, the variation observed across structures and specimens suggests that structure-level coverage reflects both autonomous planning performance and the anatomy-dependent visual accessibility of individual targets.

The current study is limited by simulation-based robot execution and a small number of anatomical specimens. The simulator does not fully reproduce tissue interaction, anatomical deformation, registration uncertainty, or robot calibration errors encountered during physical navigation, and current planning requires tens of seconds per examination trajectory. Future work will therefore focus on physical cadaveric validation, angled endoscopes such as a $30^{\circ}$ scope, larger anatomical cohorts, faster replanning, and safety-constrained control under registration and interaction uncertainty.

\section{CONCLUSION}

This work presented EndoNav, a patient-specific semantic-to-geometric grounding framework that connects anatomy-level surgeon intent to autonomous endoscopic visualization. The framework translates high-level anatomical commands into structured visualization objectives, generates target viewpoints, and realizes them through geometry-constrained endoscope motion planning. In autonomous sinus examination, EndoNav reproduced substantial portions of the anatomical surfaces observed during tracked cadaveric examinations, with robot--surgeon visualization agreement comparable to the inter-surgeon agreement observed in the experimental specimen. Evaluation across three anatomical models further demonstrated consistent examination generation across patient-specific geometries while revealing structure- and anatomy-dependent variation in achievable coverage.

These results demonstrate the feasibility of bridging semantic surgeon intent and embodied endoscope navigation through patient-specific anatomical grounding, providing a foundation for surgeon-directed robotic visualization assistants that translate clinically meaningful communication into anatomically grounded endoscopic motion.
\bibliography{bibliography}
\bibliographystyle{IEEEtran}

\end{document}